\documentclass[10pt,twocolumn,letterpaper]{article}

\usepackage{cvpr}              

\usepackage{graphicx}
\usepackage{amsmath}
\usepackage{amssymb}
\usepackage{booktabs}
\usepackage{multirow}

\usepackage[pagebackref,breaklinks,colorlinks]{hyperref}

\usepackage[capitalize]{cleveref}
\crefname{section}{Sec.}{Secs.}
\Crefname{section}{Section}{Sections}
\Crefname{table}{Table}{Tables}
\crefname{table}{Tab.}{Tabs.}

\def\cvprPaperID{4982} 
\def\confName{CVPR}
\def\confYear{2023}

\begin{document}

\title{Semi-supervised Fashion Compatibility Prediction \\by Color Distortion Prediction}

\author{Ling Xiao and Toshihiko Yamasaki \\
Department of Information and Communication Engineering, \\
Graduate School of Information Science and Technology, \\
The University of Tokyo, Tokyo, Japan 
}
\maketitle

\begin{abstract}
   Supervised learning methods have been suffering from the fact that a large-scale labeled dataset is mandatory, which is difficult to obtain. This has been a more significant issue for fashion compatibility prediction because {\it compatibility} aims to capture people's perception of aesthetics, which are sparse and changing. Thus, the labeled dataset may become outdated quickly due to fast fashion. Moreover, labeling the dataset always needs some expert knowledge; at least they should have a good sense of aesthetics. However, there are limited self/semi-supervised learning techniques in this field. In this paper, we propose a general color distortion prediction task forcing the baseline to recognize low-level image information to learn more discriminative representation for fashion compatibility prediction. Specifically, we first propose to distort the image by adjusting the image color balance, contrast, sharpness, and brightness. Then, we propose adding Gaussian noise to the distorted image before passing them to the convolutional neural network (CNN) backbone to learn a probability distribution over all possible distortions. The proposed pretext task is adopted in the state-of-the-art methods in fashion compatibility and shows its effectiveness in improving these methods' ability in extracting better feature representations. Applying the proposed pretext task to the baseline can consistently outperform the original baseline. 
\end{abstract}

\section{Introduction}
\label{sec:intro}

Fashion compatibility prediction has attracted a lot of research attention but remains a challenging task due to its subjective and ever-changing nature~\cite{Han_ACMMM_17,Vasileva_ECCV_18,Nakamura_arxiv_18,Tan_ICCV_19,Jing_TMM_2019,Lin_CVPR_20, Liu_TMM_2020,Zhang_arxiv_20,Jing_TMM_21,Sarkar_CVPR_22,Zhou_Neur_22,Zhou_Neurocomputing_22,Sarkar_CVPR_22,Xiao_ICIP_22}. Most of the existing works are supervised, which requires a large labeled dataset to ensure high accuracy. However, two main difficulties may restrict the development of supervised methods. First, a sense of {\it compatibility} is evolving, which means some labeled outfits may become out-of-date quickly. Second, labeling is really difficult and troublesome. Labeling the dataset always need some expert knowledge, at least they should have a good sense of aesthetics. In particular, fashion compatibility is based on subjective evaluation, and therefore labels are not always obvious nor consistent among the evaluators.

Although self/semi-supervised learning methods have been proposed for computer vision tasks~\cite{Noroozi_ECCV_16, Gidaris_ICLR_18,Wu_CVPR_18, Chen_ICML_20, He_CVPR_20}, they are not applied to the fashion compatibility prediction problem very much~\cite{Kim_ICCV_21,REVANUR_RecSys_21} because fashion compatibility prediction is very different from conventional object classification and recognition. 
For example, a dress can be compatible with a necklace even when they have different shapes, colors, and texture features. Fashion compatibility prediction
requires reasoning about the items' compatibility from multiple perspectives such as color, texture, pattern, and style. Specifically, it sometimes requires “complementary" compatibility, which is different from similarity. 


In this work, we propose a general pretext task for improving the baselines in fashion compatibility prediction. The core intuition behind our work is that low-level image features play key roles in people's aesthetic evaluations, if a model can't recognize the difference between the low-level information, it is impossible for the model to extract high-level compatibility information from the image. Concretely, we first generate four distorted images by adjusting the image color balance, contrast, sharpness, and brightness. Then, the distorted images are processed by adding a Gaussian noise before they are passed into a CNN backbone for a probability distribution over all possible distortions. The proposed pretext task is applied to the state-of-the-art models and its effectiveness is experimentally verified. 

Our main contributions are summarized below:
\begin{itemize}
  \item [1)]
  This paper proposes a general color distortion prediction task to force the baseline to recognize the low-level image feature to extract more discriminative representations for fashion compatibility prediction.
  \item [2)]
  We are the first to combine image distortion and Gaussian noise to make a more effective color distortion prediction task for fashion compatibility prediction.
  \item [3)]
   The proposed pretext task is applied to the state-of-the-art baselines in the fashion compatibility field and shows its effectiveness in improving their performance. It can be an inspiration for future work in self/semi-supervised learning, fashion compatibility, fashion recommendation, and other fashion-related tasks.
\end{itemize}

\begin{figure*}[t!]
  \centering
 \includegraphics[width=0.98\textwidth]{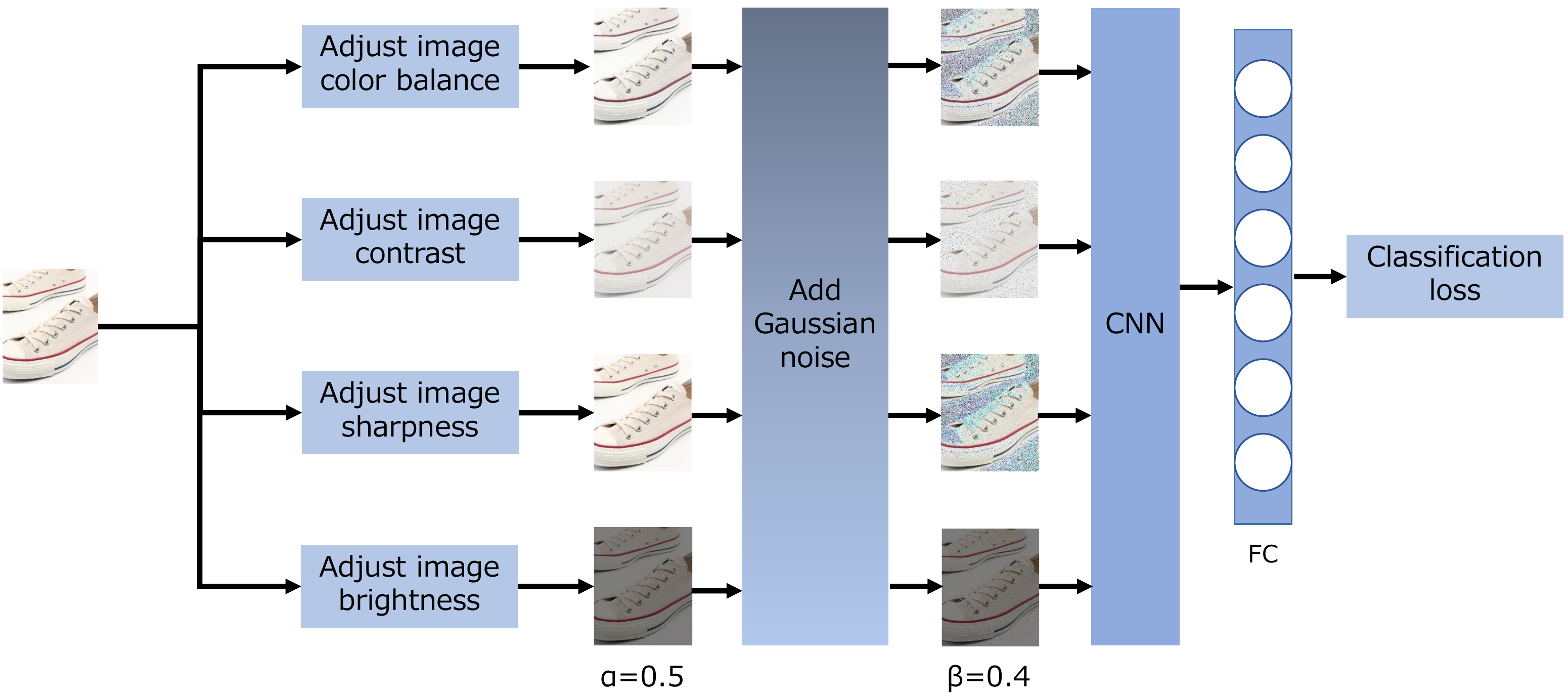}
\caption{Overview of our proposed color distortion prediction task. Here, $\alpha$ and $\beta$ are set to 0.5 and 0.4 respectively to show an example. Actually, $\alpha$ and $\beta$ are randomly selected numbers.}
   \label{fig:model}
\end{figure*}

\section{Related work}
\subsection{Self-supervised learning}

Self-supervised learning has gained much popularity across a variety of modalities because of its ability to avoid the cost of annotating large-scale datasets~\cite{Zhai_ICCV_19,Misra_CVPR_20,Liu_TKDE_21,Ziegler_CVPR_22,Lee_AAAI_22}, including image~\cite{Chen_NeurIPS_20,Grill_NeurIPS_20,Caron_NeurIPS_20}, video~\cite{Xu_CVPR_19,Qian_CVPR_21}, speech~\cite{Baevski_NIPS_20}, text~\cite{Devlin_arxiv_18}, and graphs~\cite{Velickovic_ICLR_19}. It utilizes unlabeled data to learn the underlying representations. By proposing multiple pretext tasks which use pseudo labels generated automatically
based on the attributes found in the data, models can learn good representations for multiple computer vision tasks. Handcrafted pretext tasks such as predicting rotations~\cite{Gidaris_ICLR_18}, solving jigsaw puzzles~\cite{Noroozi_ECCV_16}, and colorizing grey-scale images~\cite{Zhang_ECCV_16} provide useful
features for object recognition and detection tasks. Wu {\it et al.}~\cite{Wu_CVPR_18} proposed an Instance Discrimination (ID) pretext
task with the contrastive loss~\cite{Hadsell_CVPR_06}. ID treats each image instance
as a distinct class of its own and trains a classifier to distinguish between individual instance classes. While ID
is effective at learning strong visual representations, ID can
be biased to the texture or colors of an object which is harmful
to objection recognition. In later work, ID with strong data
augmentation techniques such as color distortion (e.g., color
jittering and gray-scale images)~\cite{Chen_ICML_20,Chen_arxiv_20} significantly improved the recognition or detection performance by providing color
and texture invariant features. Most recently, Wang {\it et al.}~\cite{Wang_CVPR_21} discovered that the batch-wise and cross-view comparisons greatly improve the positive/negative sample ratio for achieving more invariant mapping. 

However, self-supervised learning in the fashion field is not much investigated, especially in fashion compatibility prediction. To the best of our knowledge, very limited related methods are proposed ~\cite{Kim_ICCV_21,REVANUR_RecSys_21,Guan_Trans-Image-Process_22}. Kim {\it et al.}~\cite{Kim_ICCV_21} designed three pretext tasks to help the learning of fashion compatibility, including predicting color histograms, discriminating shapeless local patches, and discriminating textures from each instance. Revanur {\it et al.}~\cite{REVANUR_RecSys_21} applied random transformations to each image in the batch for shape and appearance and then measure the discrepancy between the representations of the original and perturbed images. However, these researches are not enough for the fashion field which is in high need of self-supervised learning to tackle the problems brought by the ever-changing nature of fashion. In this paper, we aim to propose a general pretext task that can be adopted to improve the baselines in the fashion compatibility field.

\subsection{Fashion compatibility prediction}

Existing work for fashion compatibility prediction can be roughly divided into supervised methods~\cite{Han_ACMMM_17,Vasileva_ECCV_18,Nakamura_arxiv_18,Tan_ICCV_19,Jing_TMM_2019,Lin_CVPR_20,Liu_TMM_2020,Zhang_arxiv_20,Jing_TMM_21,Reed_WACV_22,Xiao_ICIP_22} and self/semi-supervised methods~\cite{Kim_ICCV_21,REVANUR_RecSys_21,Guan_Trans-Image-Process_22}. Supervised methods are mainly composed of either conditional similarity networks~\cite{Vasileva_ECCV_18,Tan_ICCV_19,Jing_TMM_2019,Lin_CVPR_20}, graph neural networks~\cite{Liu_TMM_2020,Zhang_arxiv_20,Su_ACMMM_21}, or long short term memory (LSTM) based methods~\cite{Han_ACMMM_17,Nakamura_arxiv_18}. Though supervised methods are continually improved, they require a large-scale labeled dataset, which is difficult to obtain. Besides, the labeling of a fashion outfit requires expert knowledge, the notion of {\it compatibility} is ever-changing, and the labels may contain some noise. To solve these problems, some self/semi-supervised methods were proposed~\cite{Kim_ICCV_21,REVANUR_RecSys_21}. However, existing self/semi-supervised methods are far from enough, and more related research needs to be conducted to solve the problems caused by the ever-changing nature of fashion. In this paper, we aim to go further in self-supervised learning in the fashion compatibility field. This paper proposes to design a pretext task that can improve the baseline to achieve higher accuracy in fashion compatibility prediction.

\section{Methods}

With the knowledge that color, contrast, sharpness, and brightness are four key factors that can affect people's aesthetic evaluation of a cloth image. Theoretically, a feature extractor that can recognize the difference among this low-level information can help the fashion compatibility prediction. However, most of the existing methods adopted the pre-trained CNN backbone as a feature extractor, while they are pre-trained for classification and detection tasks. The two tasks focus more on the high-level general feature of a specific class. To solve this problem, this paper proposes a prediction task that can force the baselines to focus more on low-level information.

First, we generate four distorted images $\{I^{color}, I^{contrast}, I^{sharpness}, I^{brightness}\}$. Then, compatible appearance contains two kinds: complementary appearance and similar appearance. To force the feature extractor to learn complementary information, we add Gaussian noise to the distorted images to make more distortions. To force the feature extractor to learn similarity information, we define a hyper-parameter to control the distortion and set it as a randomly selected number. Figure~\ref{fig:model} shows the details of the proposed color distortion prediction task. Specifically, we define four distortions $D = \{d(_{\cdot}|y)\}_{y=1}^{4}$, where $d(_{\cdot}|y)$ is the distortion that applies to image $I$ and yields the transformed image $I^{y}=g(I|y)$ with label $y$. The CNN backbone $F(_{\cdot})$ gets an image $I^{y^{\ast}}$ as input (where the label $y^{\ast}$ is unknown to model $F(_{\cdot})$) and yields a probability distribution over all possible distortions.

\begin{equation}
  F(I^{y^{*}}|\theta) = \{F^{y}(I^{y^{\ast}}|\theta)\}_{y=1}^4,
    \label{loss1}
\end{equation}
where $F(I^{y^{\ast}}|\theta)$ is the predicted probability for the distortion with label $y$, and $\theta$ are learned parameters of model $F(_{\cdot})$.

Therefore, given a set of $N$ training images $D=\{I_{i}\}_{i=0}^{N}$, the objective that the model learns to solve is:

\begin{equation}
  min_{\theta}\frac{1}{N}\sum_{i=0}^{N}loss(I_{i}, \theta),
    \label{loss2}
\end{equation}
where the loss function $loss(_{\cdot})$ is defined as:

\begin{equation}
  loss(I_{i},\theta) = -\frac{1}{4}\sum_{y=1}^{4}\log(F^{y}(g(I_{i}|y)|\theta)).
    \label{loss3}
\end{equation}

We give details of the four image distortion designs and how to add noise to the distorted image below.

\textbf{Adjust image color balance}
Given an image $I$, to adjust the image color balance, we first transform $I$ from RGB space to gray space, and $I_{gray}$ is obtained. Then, we mix $I_{gray}$ and $I$ with a hype-parameter $\alpha_{1}$, as shown in Equation~\ref{eq:color_distortion}.

\begin{equation}
\begin{split}
  I_{gray} \!&=\! R \times 0.299 + G \times 0.587 + B \times 0.114, \\
  I_{color} \!&=\! (1.0 - \alpha_{1})I_{gray} + \alpha_{1} I.
  \end{split}
   \label{eq:color_distortion}
\end{equation}

\textbf{Adjust image contrast}
To adjust the image contrast, we first calculate the mean value of image $I$, and $I_{mean}$ is obtained. Then, we mix $I_{mean}$ and $I$ with a hype-parameter $\alpha_{2}$, as shown in Equation~\ref{eq:contrast_distortion}.

\begin{equation}
\begin{split}
 I_{gray} \!&=\! R \times 0.299 + G \times 0.587 + B \times 0.114, \\
I_{mean} &= mean(I_{gray})+0.5, \\
I_{contrast} &= (1.0 - \alpha_{2})I_{mean} + \alpha_{2} I.
\end{split}
\label{eq:contrast_distortion}
\end{equation}

\textbf{Adjust image sharpness}
To adjust the image sharpness, we first adopt a Gaussian kernel to smooth the image $I$, and $I_{smooth}$ is obtained. Then, we mix $I_{smooth}$ and $I$ with a hype-parameter $\alpha_{3}$, as shown in Equation~\ref{eq:sharpness_distortion}.

\begin{equation}
\begin{split}
  Filter &= \left(                
  \begin{array}{ccc}   
    1 & 1 & 1\\  
    1 & 5 & 1\\  
    1 & 1 & 1\\  
  \end{array}
\right),            \\
  I_{smooth} &= I* Filter, \\
   I_{sharpness} &= (1.0 - \alpha_{3})I_{smooth} + \alpha_{3} I.
\end{split}
  \label{eq:sharpness_distortion}
\end{equation}

\textbf{Adjust image brightness}
To adjust the image brightness, we first create an empty image $I_{0}$. Then, we mix $I_{0}$ and $I$ with a hype-parameter $\alpha_{4}$, as shown in Equation~\ref{eq:brightness_distortion}.

\begin{equation}
\begin{split}
  I_{0} &= zero(I), \\
  I_{brightness} &= (1.0 - \alpha_{4})I_{0} + \alpha_{4} I.
  \end{split}
   \label{eq:brightness_distortion}
\end{equation}

\textbf{Adding Gaussian noise}
We create a normal distribution $N$ with a mean of 0 and a variance of 1, as $N(0,1)$, a hype-parameter $\beta$ is assigned to control noise distortion degree. Thus the final four distorted images $I_{color}^{'}$, $I_{contrast}^{'}$, $I_{sharpness}^{'}$, and $I_{brightness}^{'}$ are given by,
\begin{equation}
\begin{split}
  I_{color}^{'} &= I_{color} + \beta N,   \\
  I_{contrast}^{'} &= I_{contrast} + \beta N,   \\
  I_{sharpness}^{'} &= I_{sharpness} + \beta N, \\
  I_{brightness}^{'} &= I_{brightness} + \beta N. \\
  \end{split}
  \label{eq:final_distortion}
\end{equation}

\begin{figure*}[t!]
\centering
\includegraphics[width=0.88\textwidth]{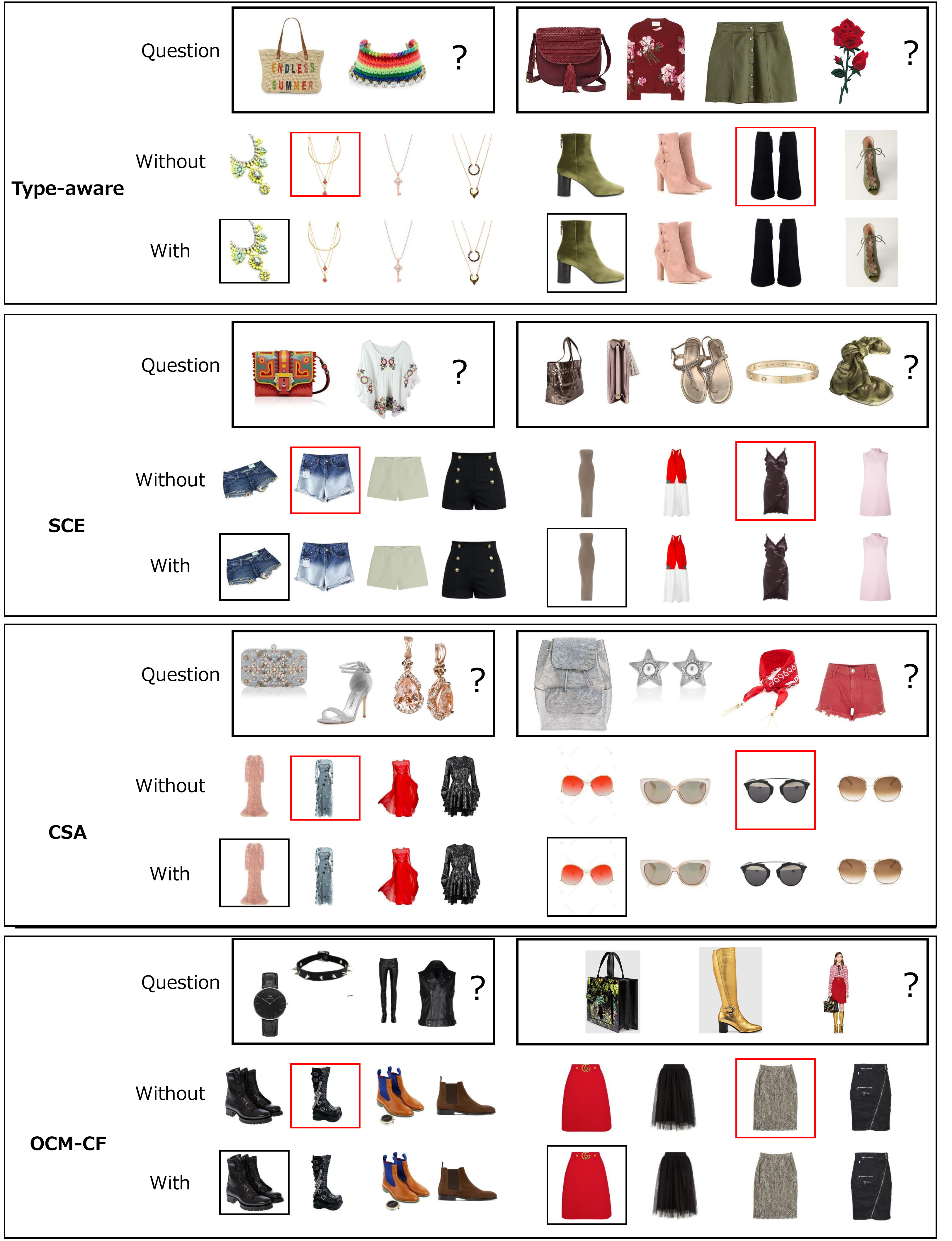}
\caption{Visualization results. Question: given incomplete outfit. Row $Without$ shows answers and the recommendation generated by the original baseline. Row $With$ shows answers and the recommendation generated by the improved baseline 2 (original baseline $+$ predicting color distortion task). The item highlighted in the black and red boxes is ground truth and false recommendation respectively. Note that, the improved baseline 2 can make an accurate recommendation from answers that share similar colors while the original baseline fails.}
\label{fig:FITB}
\end{figure*}

\begin{figure*}[t!]
\centering
\includegraphics[width=0.98\textwidth]{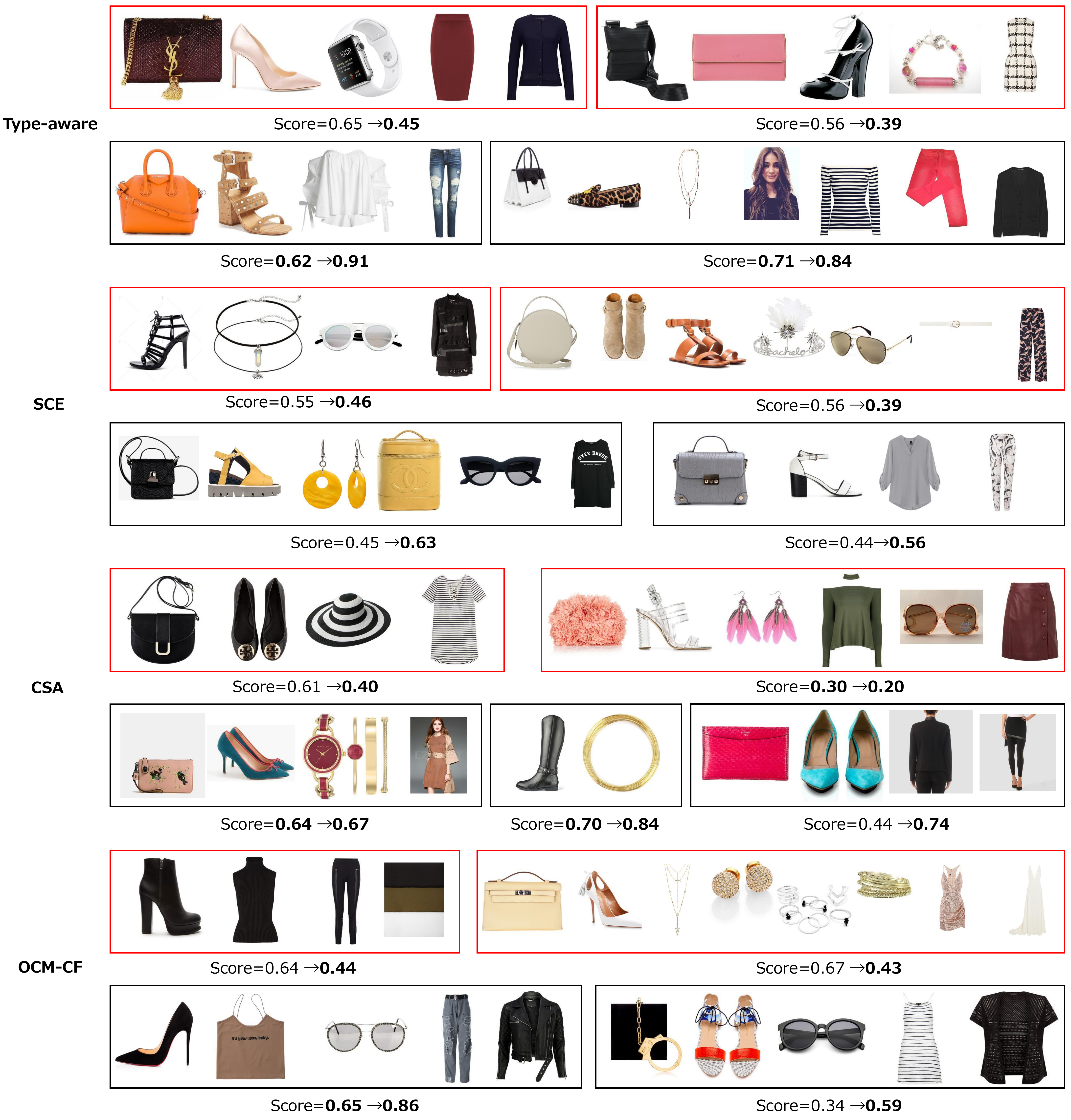}
\caption{Example results for the compatibility task. Note that, the items highlighted in the black and red boxes are labeled compatible and incompatible outfits respectively. If the score of an outfit is larger than $0.5$, it is considered compatible, and vice versa. The improved baseline 2 (original baseline $+$ predicting color distortion task) achieves a higher score for compatible outfits and a lower score for incompatible ones compared with the original baseline.}
\label{fig:Comp}
\end{figure*}

\section{Results and discussions}

\subsection{Experimental settings}

We apply the proposed pretext task to the state-of-the-art baselines and do experiments on the standard Fill-in-the-blank (FITB) and outfit compatibility tasks to evaluate the effectiveness of our pretext task. The epoch number, embedding size, margin $\mu$, and mini-batch size are 20, 32, 0.3, and 64, respectively. Since there are no ground truth negative images for each outfit, we randomly sample a set of negative images that have the same category as the positive image similar to~\cite{Vasileva_ECCV_18}. The Polyvore Outfits and Polyvore Outfits-D datasets~\cite{Vasileva_ECCV_18} are adopted. We would like to emphasize that we executed the official codes of the previous works if available (Type-aware~\cite{Vasileva_ECCV_18}, SCE-Net~\cite{Tan_ICCV_19}, and OCM-CF~\cite{Su_ACMMM_21}), re-implemented by ourselves if only part of the codes is available (CSA-Net~\cite{Lin_CVPR_20}). Therefore, the numbers of the Type-aware~\cite{Vasileva_ECCV_18}, SCE-Net~\cite{Tan_ICCV_19}, OCM-CF~\cite{Su_ACMMM_21}, and CSA-Net~\cite{Lin_CVPR_20} are slightly different from those in the original papers for a more fair comparison.

\subsection{Why image distortion?}

Table~\ref{tab:distortion_results} shows the experimental results when different $\alpha_{1-4}$ is adopted to explain why we adjust the image color balance, contrast, sharpness, and brightness to generate distorted images. Note that, we set $\alpha_{1-4}\neq 1.0$ to prevent generating the original image. The experiments are conducted by adding a predicting image distortion branch to the baseline. The numbers in bold are equal to or better than the original baseline. As we can see, even if $\alpha_{1-4}$ changes, the improved baseline 1 (original baseline $+$ predicting image distortion task) can consistently outperform the original baseline on the Polyvore Outfits and Polyvore Outfits-D datasets. Note that, we only assign random numbers to $\alpha_{1-4}$ to show this strategy is effective. People can assign other numbers to deal with their own tasks.

\begin{table*}[t!]
\caption{Experimental results when applying only the predicting image distortion task to the baseline, where different $\alpha_{1-4}$ is adopted. the improved baseline 1 (original baseline $+$ predicting image distortion task) outperforms the original baseline in most cases.}
\begin{center}
\resizebox{0.80\textwidth}{!}{
\begin{tabular}{lcccccc}
\hline
\multirow{2}{*}{Methods}& \multirow{2}{*}{Pretext-task}&\multirow{2}{*}{$\alpha_{1-4}\neq 1.0$} & \multicolumn{2}{c}{Polyvore Outfits}& \multicolumn{2}{c}{Polyvore Outfits-D}\\
\cline{4-7}
 &  & & FITB Acc. &Compat. Acc. &FITB Acc. &Compat. Acc. \\
\hline
\multirow{5}{*}{Type-aware~\cite{Vasileva_ECCV_18}} & w/o & & 54.6 &0.85
 &  54.1 & 0.83 \\
 & \multirow{4}{*}{\textbf{distortion}} & $[0.5,1.5]$ &\textbf{55.4} &\textbf{0.86}  & \textbf{54.3} &\textbf{0.83} \\ 
 & & $[0.6,1.4]$ &\textbf{55.8} & \textbf{0.85} &\textbf{55.3} & \textbf{0.83} \\
  & &$[0.7,1.3]$  &\textbf{56.0} & \textbf{0.86} &\textbf{54.5} & \textbf{0.83} \\
 &  & $[0.8,1.2]$ &\textbf{55.6} & \textbf{0.86} &\textbf{54.2} & \textbf{0.83} \\
 \hline
\multirow{5}{*}{SCE-Net~\cite{Tan_ICCV_19}} &  w/o  & &52.3 & 0.83 &  52.3& 0.82 \\
 & \multirow{4}{*}{\textbf{distortion}} & $[0.5,1.5]$  &\textbf{52.8} &\textbf{0.84}  & \textbf{52.4} &\textbf{0.82} \\ 
 & &  $[0.6,1.4]$ &\textbf{52.5} & \textbf{0.83} &\textbf{52.8} & \textbf{0.82} \\
  & & $[0.7,1.3]$ &\textbf{52.7} & \textbf{0.83} &\textbf{52.5} & \textbf{0.82} \\
 &  & $[0.8,1.2]$ &\textbf{52.6} & \textbf{0.83} &\textbf{53.0} & \textbf{0.83} \\
 \hline

\multirow{5}{*}{CSA-Net~\cite{Lin_CVPR_20}} & w/o & &54.9 &0.84 & 53.8 &0.80 \\
 & \multirow{4}{*}{\textbf{distortion}} & $[0.5,1.5]$ &\textbf{56.4} & \textbf{0.85} &\textbf{54.4} & \textbf{0.82}  \\ 
 & & $[0.6,1.4]$  &\textbf{55.3} & \textbf{0.84} &\textbf{54.9} & \textbf{0.82} \\
  & & $[0.7,1.3]$ &\textbf{56.9} & \textbf{0.86} &\textbf{54.8} & \textbf{0.82} \\
 &  & $[0.8,1.2]$ &\textbf{55.8} & \textbf{0.84} &\textbf{55.0} & \textbf{0.82} \\ \hline
  
 \multirow{5}{*}{OCM-CF~\cite{Su_ACMMM_21}}&  w/o  & &60.0 &0.81 &54.0 &0.81 \\ 
 & \multirow{4}{*}{\textbf{distortion}} & $[0.5,1.5]$ &\textbf{60.6} & \textbf{0.81} &\textbf{54.5} & \textbf{0.81}  \\ 
 & &  $[0.6,1.4]$ &59.9 & \textbf{0.81} &\textbf{54.3} & \textbf{0.81} \\
  & & $[0.7,1.3]$ &\textbf{60.2} & \textbf{0.81} &\textbf{54.4} & \textbf{0.81} \\
 &  &$[0.8,1.2]$ &\textbf{60.0} & \textbf{0.81} &\textbf{54.6} & \textbf{0.81} \\ 
\hline
\end{tabular}}
\end{center}
\label{tab:distortion_results}
\end{table*}

\subsection{Why the combination of image distortion and Gaussian noise?}

Table~\ref{tab:Main_results} shows the experimental results when the image distortion is fixed while different $\beta$ is adopted to verify the predicting color distortion task (combination of image distortion and Gaussian noise) can make a more effective pretext task. The underlined numbers are the results of the improved baseline 1 (original baseline $+$ predicting image distortion task). The numbers in bold are equal to or better than the underlined number. As we can see, even if $\beta$ changes, applying predicting color distortion task to the baseline can outperform the improved baseline 1 in most cases. Note that, we only assign random values to $\beta$ to show this strategy is effective. People can assign other numbers to deal with their own tasks.

\begin{table*}[t!]
\caption{Results when applying different pretext tasks to the baseline, where $\alpha_{1-4} \in [0.5,1.5], \alpha_{1-4}\neq 1.0$, while different $\beta$ is adopted. The predicting color distortion task outperforms the predicting image distortion task in most cases.}
\begin{center}
\resizebox{0.80\textwidth}{!}{
\begin{tabular}{lcccccc}
\hline
\multirow{2}{*}{Methods}& \multirow{2}{*}{Pretext-task}&\multirow{2}{*}{$\beta$} & \multicolumn{2}{c}{Polyvore Outfits}& \multicolumn{2}{c}{Polyvore Outfits-D}\\
\cline{4-7}
 &  & & FITB Acc. &Compat. Acc. &FITB Acc. &Compat. Acc. \\
\hline
\multirow{6}{*}{Type-aware~\cite{Vasileva_ECCV_18}} & w/o & & 54.6 &0.85
 &  54.1 & 0.83 \\
 & \textbf{distortion} & &\underline{55.4} &\underline{0.86}  & \underline{54.3} &\underline{0.83} \\ 
 & \multirow{4}{*}{\textbf{distortion+noise}} & [0.01,0.05]  &\textbf{55.9} & \textbf{0.86} &53.9 & \textbf{0.83} \\
  & & [0.02,0.05] &\textbf{56.3} & \textbf{0.86} &\textbf{54.6} & \textbf{0.83} \\
 &  & [0.03,0.05]  &\textbf{55.8} & \textbf{0.86} &53.8 & \textbf{0.83} \\
 &  & [0.04,0.05]  &\textbf{56.1} & \textbf{0.86} &\textbf{54.3} & \textbf{0.83} \\
 \hline
\multirow{6}{*}{SCE-Net~\cite{Tan_ICCV_19}} &  w/o  & &52.3 & 0.83 &  52.3& 0.82 \\
 & \textbf{distortion} & &\underline{52.8} &\underline{0.84}  & \underline{52.4} &\underline{0.82} \\  
 & \multirow{4}{*}{\textbf{distortion+noise}} & [0.01,0.05]  &\textbf{53.6} & 0.83 &\textbf{53.5} & \textbf{0.83} \\
  & & [0.02,0.05]  &\textbf{53.5} & \textbf{0.84} &\textbf{52.9} & \textbf{0.83} \\
 &  & [0.03,0.05]  &\textbf{53.1} & \textbf{0.84} &\textbf{52.7} & \textbf{0.82} \\
 &  &[0.04,0.05]  &\textbf{53.3} & \textbf{0.84} &\textbf{53.0} & \textbf{0.83} \\
 \hline

\multirow{6}{*}{CSA-Net~\cite{Lin_CVPR_20}} & w/o & &54.9 &0.84 & 53.8 &0.80 \\
& \textbf{distortion}& &\underline{56.4} & \underline{0.85} &\underline{54.4} & \underline{0.82} \\ 
&\multirow{4}{*}{\textbf{distortion+noise}}& [0.01,0.05]  &\textbf{56.6} & \textbf{0.86} &\textbf{54.5} & \textbf{0.82}\\ 
 &  & [0.02,0.05]  &\textbf{56.7} & \textbf{0.85} &\textbf{55.0} & \textbf{0.82} \\  
  &  &[0.03,0.05]  &\textbf{57.0} & \textbf{0.86} &\textbf{55.1} & \textbf{0.82} \\ 
  &  & [0.04,0.05]  &\textbf{57.1} & \textbf{0.85} &\textbf{54.8} & \textbf{0.82} \\
  \hline
  
 \multirow{6}{*}{OCM-CF~\cite{Su_ACMMM_21}}&  w/o  & &60.0 &0.81 &54.0 &0.81 \\ 
 & \textbf{distortion} & &\underline{60.6} &\underline{0.81}  & \underline{54.5} &\underline{0.81} \\ 
 & \multirow{4}{*}{\textbf{distortion+noise}} & [0.01,0.05]  &60.4 & \textbf{0.81} &\textbf{54.6} & \textbf{0.81} \\
  & & [0.02,0.05]  &\textbf{60.9} & \textbf{0.82} &\textbf{54.5} & \textbf{0.81} \\
 &  & [0.03,0.05]  &\textbf{60.7} &0.80&\textbf{54.9} & \textbf{0.81} \\
 &  & [0.04,0.05]  &\textbf{60.8} & \textbf{0.82} &\textbf{55.3} & \textbf{0.81} \\
\hline
\end{tabular}}
\end{center}
\label{tab:Main_results}
\end{table*}

\subsection{Visualization}
Figures~\ref{fig:FITB} and~\ref{fig:Comp} show some visualization results. From Figure~\ref{fig:FITB}, for the answers that share a similar color, the improved baseline 2 (original baseline $+$ predicting color distortion task) can make accurate recommendations while the original baseline fails. Moreover, from Figure~\ref{fig:Comp}, the improved baseline 2 generates a higher score for compatible outfits and a lower score for incompatible ones compared with the original baseline. It is demonstrated that the proposed pretext task could improve the ability of the baseline when reasoning about fashion items' compatibility with the image.

\begin{figure*}[t!]
  \centering
   \includegraphics[width=0.98\linewidth]{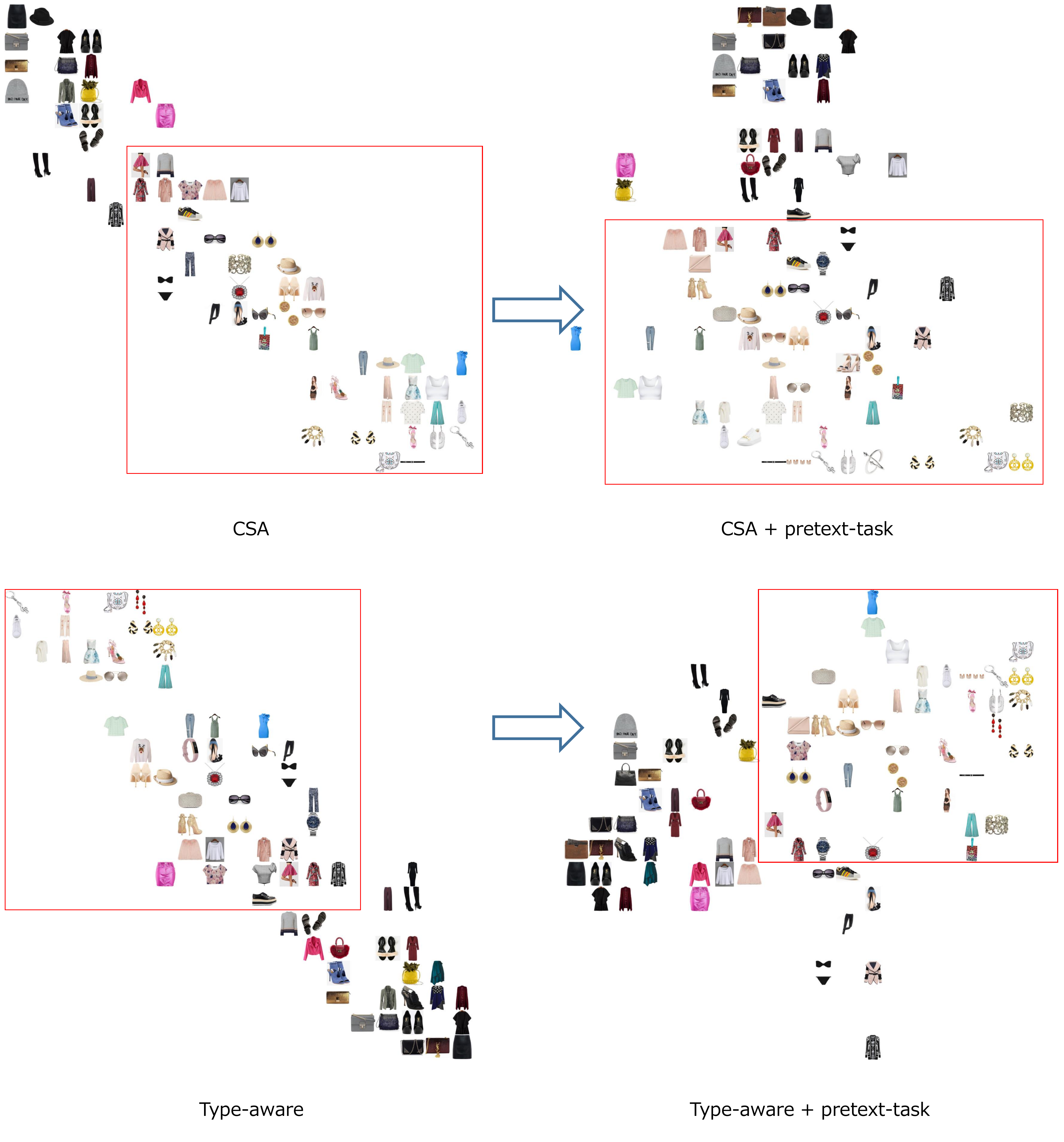}
   \caption{The t-SNE visualization of the compatibility learned using different methods. The Type-aware~\cite{Vasileva_ECCV_18} and CSA-Net~\cite{Lin_CVPR_20} are taken as examples. As you can see from the red bounding box area, the improved baseline 2 (original baseline $+$ predicting color distortion task) learns a better embedding space for fashion compatibility than the original baseline.}
   \label{fig:t-SNE}
\end{figure*}

We also visualize the t-distributed stochastic neighbor embedding (t-SNE) learned using different methods. As we can see from Figure~\ref{fig:t-SNE}, when the proposed pretext task is adopted, the model can learn a better embedding space for fashion compatibility prediction. This further demonstrates the effectiveness of the proposed method.

\subsection{Discussions}

The simple formulation of our pretext task has several advantages. There are publicly available python packages that make it easy to apply, and only several lines of code can do it. Moreover, as we can see in the experimental section of the paper when our pretext task is adopted, the baseline is improved by a big margin.

\section{Conclusions}

In this paper, we explored self-supervised pretext tasks for fashion compatibility prediction. Based on the knowledge that low-level information plays an important role when recommending a fashion item for an outfit, we proposed a general color distortion prediction task to improve the baseline's ability in reasoning the compatibility among fashion items by forcing the baseline to recognize the difference among several distortions. Specifically, we first generate four distorted images by adjusting the image color balance, contrast, sharpness, and brightness, respectively. Then, to increase the difference among distorted images, the distorted images are processed with a Gaussian noise before passing to the CNN backbone. The proposed pretext task is applied to state-of-the-art methods in the fashion compatibility field to evaluate the effectiveness of our method in improving the baselines. Experiments on the publicly available Polyvore Outfits and Polyvore Outfits-D datasets demonstrated that when our pretext task is adopted, the baseline can consistently be improved. The proposed pretext tasks can be adopted into other models in the fashion field easily and it has a high potential to help the learning of other computer vision tasks.

{\small
\bibliographystyle{ieee_fullname}
\bibliography{egbib}
}

\end{document}